\documentclass[10pt,journal,cspaper,compsoc]{IEEEtran}
\usepackage{amssymb,amsmath}
\usepackage{graphicx}
\usepackage{subcaption}
\usepackage{multirow}
\usepackage[noadjust]{cite}
\usepackage{csquotes}
\usepackage{stfloats}
\usepackage{color}

\title{Multi-label Multi-task Deep Learning\\ for Behavioral Coding}

\author{James~Gibson,
	David~C.~Atkins,
	Torrey~Creed,
	Zac~Imel,
	Panayiotis~Georgiou,
        Shrikanth Narayanan
\IEEEcompsocitemizethanks{\IEEEcompsocthanksitem James Gibson, Panayiotis Georgiou, Shrikanth Narayanan are with the Department of Electrical Engineering, University of Southern California, Los Angeles, CA 90089 USA (e-mail: jjgibson@usc.edu).
\IEEEcompsocthanksitem David C. Atkins is with the Department of Psychiatry and Behavioral Sciences, University of Washington, Seattle, WA 98195 USA.
\IEEEcompsocthanksitem Torrey~Creed is with the Department of Psychiatry, Perelman School of Medicine, University of Pennsylvania, Philadelphia, PA 19104 USA.
\IEEEcompsocthanksitem Zac~Imel is with the Department of Psychiatry, School of Medicine, University of Utah, Salt Lake City, UT 84112  USA.
}
}

\begin{document}

%



\IEEEcompsoctitleabstractindextext{%
\begin{abstract}
We propose a methodology for estimating human behaviors in psychotherapy sessions using mutli-label and multi-task learning paradigms.
We discuss the problem of behavioral coding in which data of human interactions is the annotated with labels to describe relevant human behaviors of interest.
We describe two related, yet distinct, corpora consisting of therapist client interactions in psychotherapy sessions.
We experimentally compare the proposed learning approaches for estimating behaviors of interest in these datasets.
Specifically, we compare single and multiple label learning approaches, single and multiple task learning approaches, and evaluate the performance of these approaches when incorporating turn context.
We demonstrate the prediction performance gains which can be achieved by using the proposed paradigms and discuss the insights these models provide into these complex interactions.
\end{abstract}
\begin{IEEEkeywords}
\end{IEEEkeywords}
}

\maketitle


\section{Introduction} \label{S:intro}

\IEEEPARstart{U}{nderstanding} and describing human behavior is an immensely multifaceted task.  
In conversation, participants' behaviors unfold and evolve over time, occurring in both brief and extended time scales.  
Additionally, these behaviors are often co-occurring and intrinsically related to one another.
The complexity of these interactions presents an opportunity to investigate machine learning paradigms which may better reflect the intricacies of these types of data than traditional machine learning procedures.
Specifically, we explore multi-label and multi-task learning approaches for predicting behaviors in psychotherapy.
The multi-label system is trained to predict co-occurring behaviors, at the turn or session level, of the therapists and clients during psychotherapy sessions.
Subsequently, we propose a multi-task system to learn behaviors across multiple psychotherapy domains.
Finally, we evaluate these methodologies when context across multiple turns in the interactions is incorporated.
  
In psychotherapy, the therapist seeks to work with the client to create change in cognitions, emotions, or behaviors that are causing distress or impairment.
There are a variety of behaviors employed in this process that vary according to the type of therapy, the aims of the therapy, the client's characteristics, and the training and skill of the therapist.
Researchers have suggested that for a variety of symptoms the type of therapy may not significantly affect outcomes \cite{imel2008distinctions}.
Thus, it can be assumed that there are some underlying mechanisms at work that are common across psychotherapy approaches.
In fact, research suggests that there are a number of common factors across evidence based psychotherapies, such as alliance and empathy \cite{wampold2015important}.
This has lead to the efforts to develop an evidence-based psychotherapy that is effective across many common mental health disorders \cite{farchione2012unified}.

In this work, we use two distinct, yet related, psychotherapy approaches to serve as example domains in which we evaluate automatic behavioral coding (ABC) systems employing multi-label and multi-task learning frameworks.
Motivational Interviewing (MI) is a client centered approach to therapy that aims to promote behavior change in clients by exploring and resolving ambivalence.
Cognitive Behavior Therapy (CBT), in contrast to motivational interviewing, is focused on developing coping strategies aimed at decreasing symptoms.
Both therapies are goal-oriented, evidence based practices which are concerned with client behavior change.
So, despite differences in approach, there is significant overlap in philosophical orientation and employed techniques in the domains.

\subsection{Behavioral Coding in Psychotherapy}
In psychotherapy  research, behavioral coding is the process of identifying and codifying the behaviors which are most relevant to the aims of therapy \cite{bakeman2000behavioral}.  
The objective of this procedure is to define clear and broadly applicable behavioral `codes' which represent target behavioral constructs that are of interest to a particular study or line of inquiry.  
Behavioral observation and coding is common practice in many subfields of psychology including diagnosing autism \cite{lord2012autism}, family and marital observational studies \cite{margolin1998nuts, christensen2004traditional}, and several forms of psychotherapy \cite{miller2009toward, creed2016implementation}.
Because manual behavioral coding is costly and time-intensive, there is an opportunity for the development of methodologies aimed at automating aspects of this process.

\subsection{Machine Learning for Behavioral Coding}
\label{S:mlforbc}

There have been numerous works aimed at using human interaction data to help automate and inform the behavioral observation and coding process in domains such as marital therapy \cite{black2011toward}, motivational interviewing \cite{xiao2016technology, tanana2016comparison}, and autism diagnosis \cite{bone2016use}.
Additionally, there have been many features explored in these works including acoustic and prosodic speech features \cite{black2011toward, xiao2013modeling, xiao2015analyzing}, lexical and semantic features \cite{gibson2015predicting, perez2017predicting}, automatically derived lexical features \cite{georgiou2011thats}, and visual features \cite{xiao2014power}.

While the majority of systems proposed have been traditional fully supervised learning approaches, prior to this work, we proposed Multiple Instance Learning (MIL) for behavioral coding in couples therapy \cite{gibson2017multiple}.
MIL is a semi-supervised learning paradigm, in which several samples share a single label, thus an MIL system attempts to learn a many-to-one mapping between samples and labels.
In the case of behavioral coding, this mapping is between the multiple turns (samples) to session level behavioral codes (labels).
In the present work, the proposed approaches can be thought of as a type of multiple instance learning.
However, the samples in this work are treated as sequences rather than independent observations and therefore better reflect the temporal nature of human interactions.


Using a sequential model (a Conditional Random Field) to predict sequences of utterance level behavioral codes in psychotherapy was first proposed in \cite{can2015dialog}.
This work also proposed using dialog acts as a proxy for utterance level behavioral codes and demonstrated that using dialog acts for predicting session level behaviors achieved competitive performance to using carefully defined and annotated utterance level behaviors.
The first application of neural networks to behavioral coding was proposed in \cite{tanana2015recursive}.
The authors propose a recursive neural network for deriving an utterance representation and use a Maximum Entropy Markov Model (MEMM) to perform detection of client change talk and sustain talk at the individual utterance level.
Recurrent neural networks (RNNs) were first proposed for behavioral coding in \cite{xiao2016behavioral} and \cite{gibson2016deep}.
In \cite{xiao2016behavioral}, the authors compare Long short term memory (LSTMs) and Gated Recurrent Units (GRUs), two varieties of RNNs, for predicting utterance level behavioral codes from word embeddings.
In \cite{gibson2016deep}, LSTMs are used for encoding turn context from turn embeddings for predicting utterance level behaviors which is subsequently used as the lower layers of a deeper system that predicts sessions level empathy in psychotherapy interactions.
Recently, in \cite{singlak2018using}, the authors proposed using multimodal word-level based LSTMs trained with prosodic and lexical features for predicting utterance level codes in motivational interviewing sessions.
In \cite{flemotomos2018language}, the authors compare several lexical and semantic feature representations for predicting session level behaviors in cognitive behavioral therapy sessions.

\subsection{Multi-label Learning}

Multi-label learning is a machine learning paradigm in which each sample is associated with several, possibly related, labels \cite{zhang2014review}.  
Such a framework allows for a model to learn more general features because they must be relevant to multiple targets.
Also, this approach allows a model to account for relationships between labels which can be especially useful for predicting less frequent labels.
Multi-label learning has been explored for a wide variety of applications including functional genomics, text categorization, and scene classification \cite{zhang2014review, sorower2010literature} and for a variety of classifiers including K-nearest neighbors \cite{zhang2007ml}, support vector machines \cite{elisseeff2002kernel}, and deep neural networks \cite{zhang2006multilabel}.

In behavioral coding, each code attempts to capture a distinct behavior of interest.
However many of these behaviors are fundamentally related; for example open questioning and reflective listening are considered the skills of a well trained motivational interviewer, whereas confrontations are not.
In this sense behavioral coding is a problem with multiple interrelated outputs which motivates investigating a multi-label learning approach.


\subsection{Multi-task Learning}

Multi-task learning is a machine learning paradigm in which a single model is optimized for more than one task \cite{caruana1998multitask}.  
Such a model can share part or all of its architecture save for the outputs which are dedicated to specific tasks.
These tasks are often related, allowing the model to key in on features of general importance.
This approach also allows for a model to experience more data even though the labels of each sample may not be available for all tasks.
This framework has shown success in a variety of domains including text categorization \cite{lan2017multi}, head pose estimation \cite{yan2016multi}, emotion recognition \cite{pons2018multi}, and distance speech recognition  \cite{zhang2017attention}.
Recently, Liu et al., have proposed an adversarial training approach for multi-task networks using a network consisting of shared and private layers where the an adversarial loss is used to force the shared layers to learn task invariant features \cite{liu2017adversarial}.

Data from many behavioral coding domains are of a sensitive and private nature.
For this reason it is often difficult to obtain such data.
Thus paradigms like multi-task learning which can learn shared representations across related domains allow for inclusion of data from corpora, even if these corpora do not share identical types of interaction and behavioral coding schemes.

\subsection{Multi-resolution Learning}
\label{S:contextlearning}

%

Multi-resolution learning attempts to take advantage of hierarchies existing in data or labels.
For example, in the case of document classification, each document is comprised of multiple sentences.
In this case, the representation can be formulated with layers that learn a mapping from word to sentence followed by layers responsible for learning a mapping from sentence to document label \cite{tang2015document, yang2016hierarchical}.
In addition to document classification, other notable application of hierarchical learning include sequence generation \cite{li2015hierarchical}, image classification \cite{katole2015hierarchical}, and sentiment analysis \cite{lakkaraju2014aspect}.

With respect to human behavioral coding, there are many resolutions at which these interactions can be evaluated, including sessions which are comprised of speaker turns which are in turn comprised of speaker verbal and non-verbal expressions as well as the behaviors which are expressed and at times coded at the utterance and session levels.
Therefore it is important to incorporate contextual information, whether across words or turns, to learn representations which reflect the nature of these interactions.

\section{Methodology}

In this work, we employ deep learning architectures as a means of comparing single/multi-label and single/multi-task learning paradigms.
In table \ref{T:notation}, we present an reference for the notation used throughout the paper.

\begin{table}[thb]
\caption{\label{T:notation} {Notation reference.}}
\vspace{2mm}
\centerline{
\begin{tabular}{|ll|}
\hline
Symbol & Meaning \\
\hline \hline
$i$ & session index \\
$j$ & turn index \\
$k$ & word index \\
$N$ & number of sessions \\
$M_i$ & number of turns in session \\
$K_{ij}$ & number of words in turn \\
$L$ & number of labels in multi-label set \\
$C$ & turn context \\ 
$\textrm{w}_{ij}$ & turn word sequence \\
$w_{ijk} $ & word embedding vector \\
$h_{ijk}$ & hidden state of word encoder \\
$X_{ij}$ & turn vector representation \\
$r_{ij}$ & speaker role \\
$X^C_{ij}$ & set of turn context vectors \\
$g_{ij}$ & hidden state of turn encoder \\
$V_{ij}$ & turn context vector representation \\
$y_{ij}$ & turn label \\
$Y_{ij}$ & turn multi-label set \\
$z_i$ & session label \\
$Z_i$ & session multi-label set \\
$s_{ij} $ & sample weight \\
$T_{ij}$ & task relevance of the ${ij}^{\text{th}}$ sample \\
\hline
\end{tabular}
}
\end{table}

\begin{figure}[!hb]
\begin{center}
  \begin{subfigure}[t]{.45\columnwidth}
    \includegraphics[width=\columnwidth]{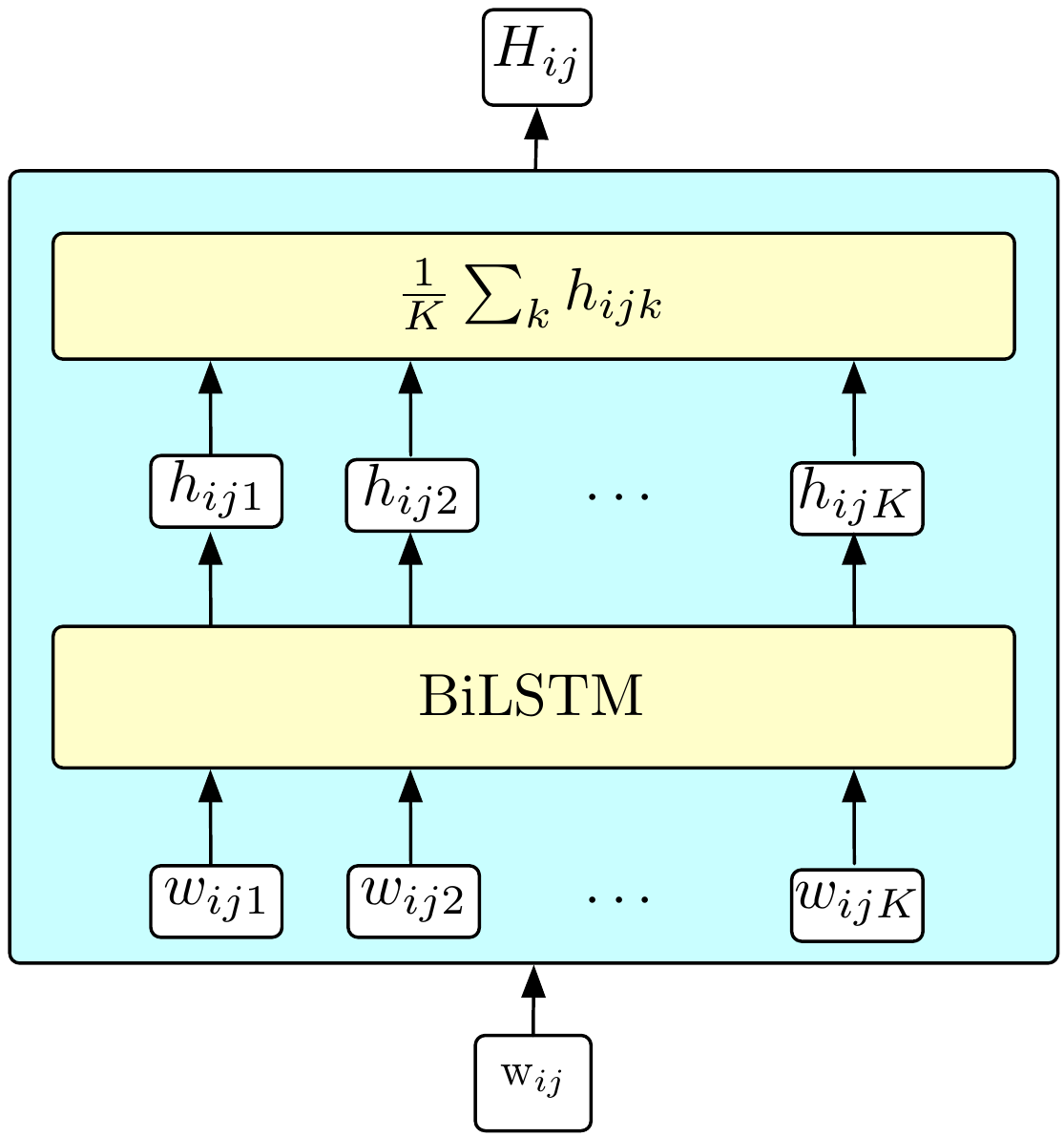}
    \caption{Word encoder}
    \label{F:word_encoder}
  \end{subfigure}
  \begin{subfigure}[t]{.45\columnwidth}
    \includegraphics[width=\columnwidth]{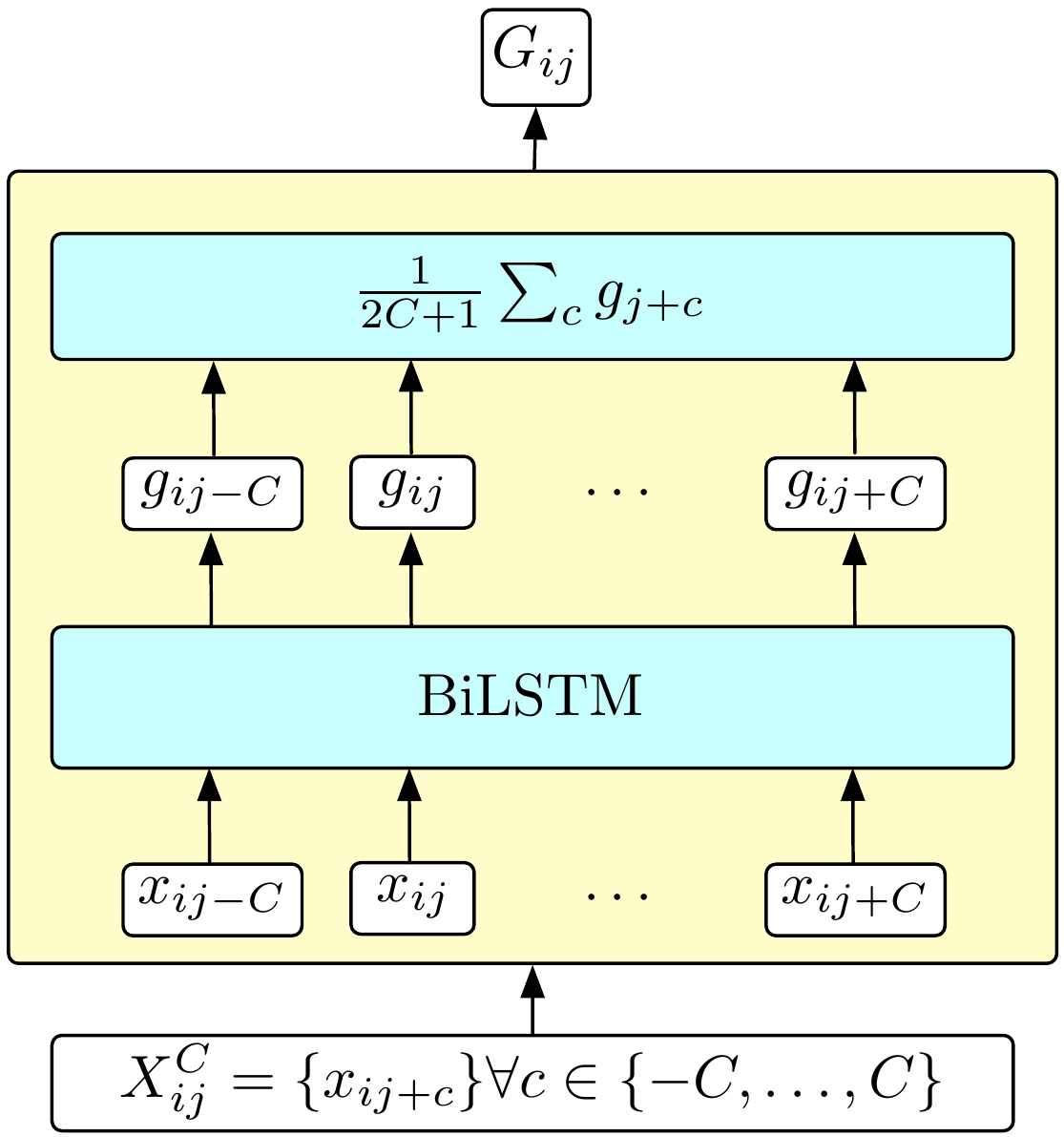}
    \caption{Turn encoder}
    \label{F:turn_encoder}
  \end{subfigure}
    \begin{subfigure}[c]{.44\columnwidth}
    \includegraphics[width=\columnwidth]{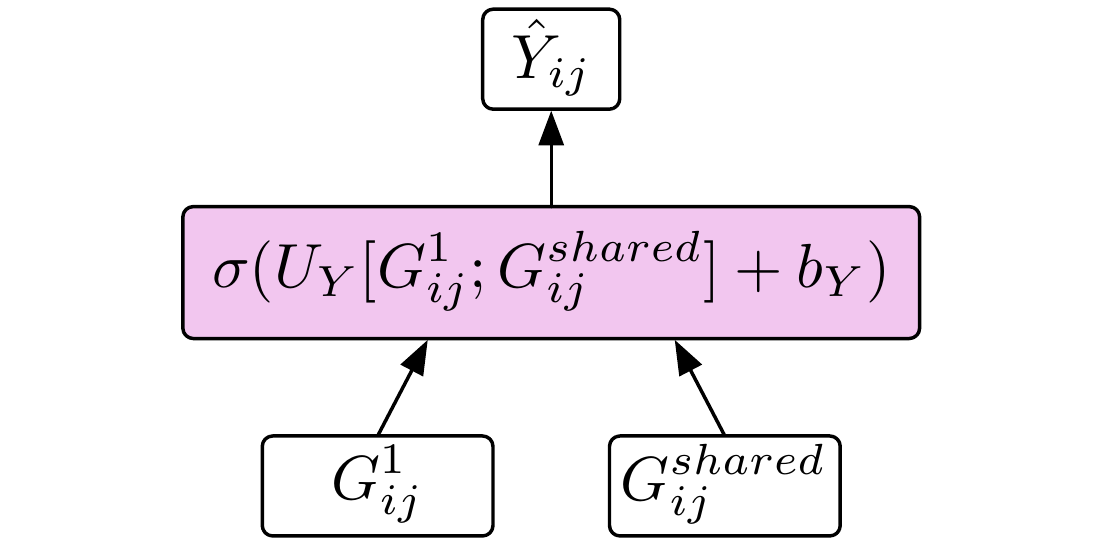}
    \caption{Turn level predictor}
    \label{F:turn_predictor}
  \end{subfigure}
    \begin{subfigure}[c]{.44\columnwidth}
    \includegraphics[width=\columnwidth]{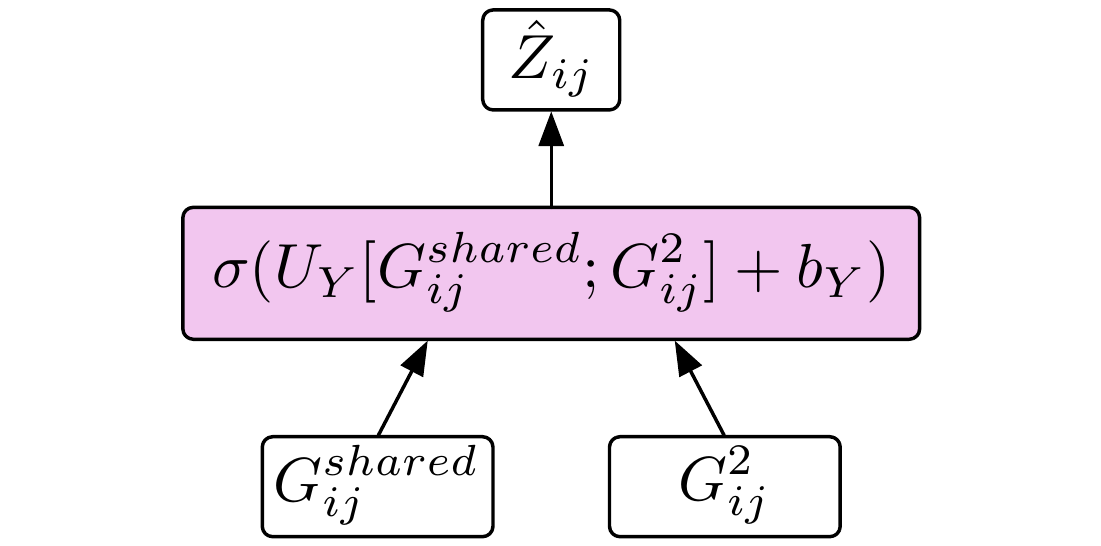}
    \caption{Session level predictor}
    \label{F:session_predictor}
  \end{subfigure}
    \begin{subfigure}[c]{.44\columnwidth}
    \includegraphics[width=\columnwidth]{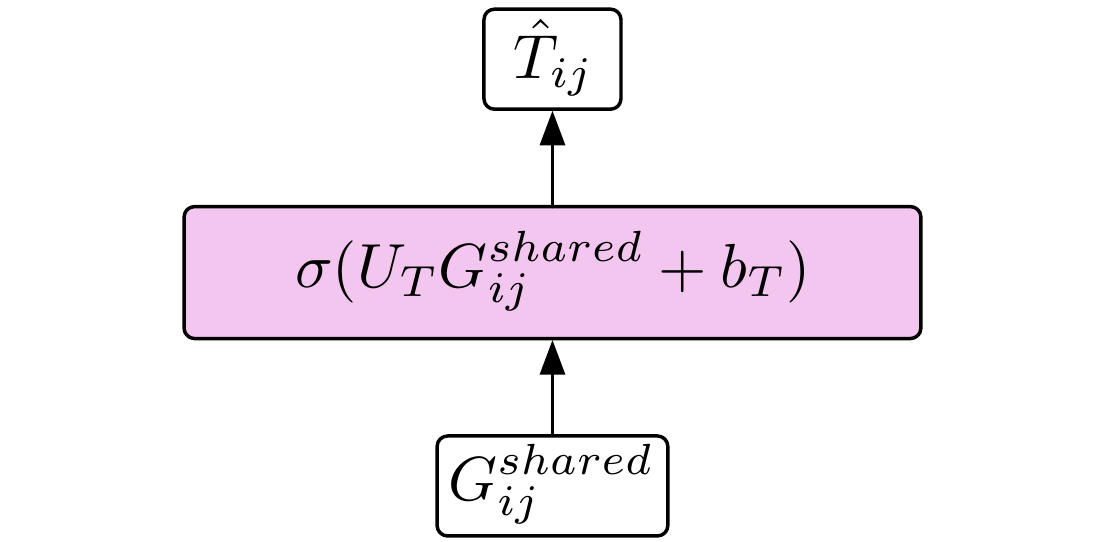}
    \caption{Task discriminator}
    \label{F:task_discriminator}
  \end{subfigure}
  \caption{Diagram of encoding and prediction networks}\label{F:lstms}
\end{center}
\end{figure}

In this work, we assume the $i^\text{th}$ session, $i \in \{1,2,...,N\}$, to be comprised of a series of turns, $j \in \{1,2,...,M_i\}$, which are comprised of word sequences, $W_{ij}$.
A turn is defined as all words which are spoken in a speaker homogenous region, the speaker having a particular role, $r_{ij}$, as either the therapist or client in our expository domain.
Each word sequence is represented by a series of word embedding vectors, $\text{w}_{ij} = \{w_{ij1},w_{ij2},...,w_{ijK_{ij}}\}$.
The word sequences are input to the word encoder, consisting of a bidirectional long short term memory (BiLSTM) network \cite{hochreiter1997long}.
The resulting hidden states of the BiLSTM are averaged, giving a vector representation of the $j^\text{th}$ turn, $x_{ij}$.
A visualization of the word encoder is given in figure \ref{F:word_encoder}.

\begin{figure*}[htbp]
\begin{center}
\includegraphics[width=2\columnwidth]{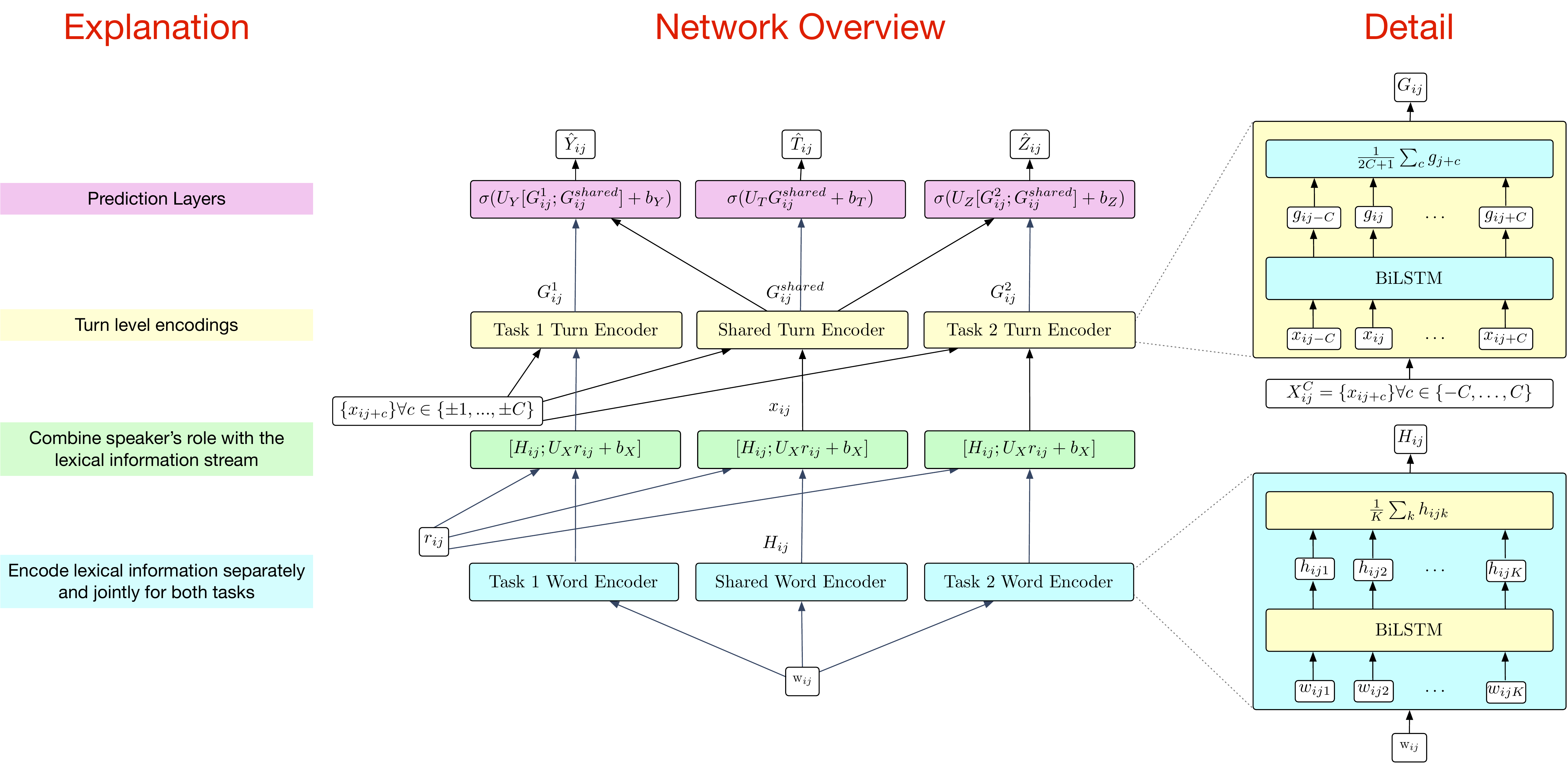}
\caption{Multi-task System Overview}
\label{F:system_overview}
\end{center}
\end{figure*}

\subsection{Multi-label Learning for Behavioral Coding}

Behavioral observation and coding can be applied at a variety of temporal granularities, including at the utterance, turn, and session levels.  
The behavioral codes, or labels, annotated in a particular segment are often co-occurring and related.
Thus, viewing these individual labels as a set of relevant labels, allows for casting the behavioral coding problem as a multi-label learning problem.
In this multi-label learning scenario, a sample, e.g., $\text{w}_{ij}$, has an associate set of labels, $Y_{ij}$, where $Y_{ij}(l)=1$ if the $l^\text{th}$ label is true for that sample and $Y_{ij}(l)=0$, otherwise.  For session level labels, the posterior of estimated labels from the predictor is averaged across turns in each session:
\begin{equation}
\hat{Z}_i = \frac{1}{M_i} \sum_{j=1}^{M_i} \hat{Z}_{ij},
\end{equation}
for the session level prediction.

\subsubsection{Multi-label Learning with Deep Neural Networks}

Deep neural networks provide a flexible architecture for multi-input and multi-output learning paradigms.  
A multi-output network can be interpreted as a multi-label network when the network weights are fully shared by the multi-label outputs.  



For the multi-label loss we use binary cross entropy loss, summed across the multi-label outputs.  This loss does not explicitly take into account correlations between the labels but rather relies on the shared network weights to encode this information.  The multi-label binary cross entropy loss is given by:
\begin{equation} \label{E:binary_crossentropy}
\begin{aligned}
E = & - \sum_{i=1}^N \sum_{j=1}^{M_i} \sum_{l=1}^L Y_{ij}(l)\cdot \log\left(\hat{Y}_{ij}(l)\right)  \\ &+ \left(1 - Y_{ij}(l)\right) \cdot \log\left(1 - \hat{Y}_{ij}(l)\right).
\end{aligned}
\end{equation}

\subsubsection{Multi-label Sample Weights}

Class imbalance is a common problem in machine learning that can drastically impact model training and generalization.
A common approach is to weigh the loss function so losses incurred by samples of less frequent classes are weighted more heavily, so as to increase the impact of those samples in the model.
In multi-label problems class imbalance is no longer clearly defined as labels are no longer individual but parts of a multi-label set.
One option would be to weigh the loss function according to the power set of the multi-label set, $2^\mathbf{Y}$.
However, because the power set grows exponentially with the number of labels the label co-occurrence distribution becomes sparse which can lead to overfitting.
In \cite{zhang2015towards}, the authors propose an algorithm that seeks to jointly learn binary class learning for each label and multi-class learners for first order pairs of labels to create a predictive multi-label model.
While, this approach has shown promise for addressing class imbalance in multi-label datasets, it does require learning additional parameters which is undesirable in deep learning settings where the number of parameters to be learned is typically already high.
Instead, we propose a heuristic approach that weighs the loss function according to the average frequency of the labels appearing in a given label set.
Each label contributes a weight according to the inverse frequency of that individual label, i.e.,
\begin{equation}
s_{ij}(l) = \begin{cases} \frac{\sum_{ij} 1-Y_{ij}(l)}{\sum_{ij} Y_{ij}(l)}, & \text{if} \thickspace Y_{ij}(l)=1. \\
				1, & \text{if} \thickspace Y_{ij}(l)=0.
	       \end{cases}
\end{equation}
The mean of these weights is taken as the multi-label sample weight:
\begin{equation}
s_{ij} = \frac{1}{L}\sum_{l=1}^{L} s_{ij}(l). 
\end{equation}
Thus when a given loss is computed for that sample it is weighted according to:
\begin{equation}
E = \sum_{i=1}^N \sum_{j=1}^{M_i} s_{ij} \text{Loss} (Y_{ij}, \hat{Y}_{ij}).
\end{equation}


\subsection{Multi-task Learning of Behavioral Codes}
\label{S:mtlmethod}

We show an overview of the proposed single-task and multi-task learning systems in figure \ref{F:system_overview}.  
Our multi-task system follows an adversarial approach proposed by \cite{liu2017adversarial}.  
This system consists of word and turn level encoders, shown in Figures \ref{F:word_encoder} and \ref{F:turn_encoder},  for each individual task as well as shared encoders to jointly encode information from both tasks.  
The output of these encoders is then concatenated and fed to a predictor, shown in Figure \ref{F:turn_predictor} for each task as well as a shared predictor, shown in Figure \ref{F:task_discriminator} , that attempts to discriminate between the tasks.  
The gradient from the task discriminator is reversed to the shared encoder in order to make the shared encoder task invariant.  
Additionally, orthogonality constraints are placed on the encoder outputs between the task specific and shared encoders in order to ensure that they are not encoding redundant representations.  
The total multi-task loss is computed as:
\begin{equation}
E_{\text{total}} = \sum_m E_m + \lambda E_{\text{task}} + \gamma E_{\text{diff}},
\end{equation}
where $E_m$ is the loss of the $m^{\text{th}}$ task, $E_{\text{task}}$ is the loss of the task discriminator, $E_{\text{diff}}$ is the loss of the orthogonality constraint, and $\lambda$ and $\gamma$ are hyper-parameters for weighting the respective losses.
The task discriminator attempts to predict which domain a particular sample belongs to.
It is a single feed-forward layer with a sigmoid activation, i.e.,
\begin{equation}
\hat{T}_{ij} = \sigma(U_T G_{ij}^{\text{shared}} + b_T).
\end{equation}
The task discriminator loss is the binary cross entropy between the reference task label, $T_{ij}$, and the prediction from the task discriminator $\hat{T}_{ij}$, i.e.,
\begin{equation} \label{E:task_loss}
\begin{aligned}
E_{\textrm{task}} = & - \sum_{i=1}^N \sum_{j=1}^{M_i} \sum_{l=1}^L T_{ij}(l)\cdot \log\left(\hat{T}_{ij}(l)\right)  \\ &+ \left(1 - T_{ij}(l)\right) \cdot \log\left(1 - \hat{T}_{ij}(l)\right).
\end{aligned}
\end{equation}
The gradient from the task discriminator is reversed to the shared encoder making it increasingly difficult to predict the task as training proceeds thus resulting in a task invariant representation in the shared encoder.
The orthogonality constraint loss is given by:
\begin{equation}
E_{\text{diff}} = \sum_m || (G^{\text{shared}})^{\text{T}} G^{m} || ^2,
\end{equation}
where $G^\text{shared}$ is the output of the shared encoder, $G^{m}$ is the result of the $m^\text{th}$ task encoder, and $||\cdot||^2$ is the squared Frobenius norm.


\subsection{Learning with Turn Context}
Because the context of a turn, i.e., the surrounding turns, can provide important discriminative information for predicting the label of a particular turn, the word encoder is followed by a turn encoder.
The turn encoder is provided both the vector, $H_{ij}$, the result of the word encoder, as well as the role of the speaker of the turn, $r_{ij}$ (i.e., an indicator variable of whether the speaker is the therapist or client). 
The speaker role undergoes a single linear transformation layer whose output is of equal size to the number of labels $L$.
These two variable are then concatenated into a vector representation of the turn, i.e.,
\begin{equation}
X_{ij} = [H_{ij} ; U_X r_{ij} + b_X].
\end{equation}
The current turn representation, $X_{ij}$, is then concatenated with the $C$ preceding and following turns, $X^C_{ij} = \{X_{ij-C}, X_{ij-C+1},..., X_{ij},...,X_{ij+C-1}, X_{ij+C}\}$ and input a BiLSTM which will encode the turn context representations.
The hidden states of the BiLSTM are averaged resulting in a turn context vector representation, $G_{ij}$.
A visualization of the turn encoder is given in figure \ref{F:turn_encoder}.

%

\section{Data}

\subsection{Motivational Interviewing Corpus}

The Motivational Interviewing Corpus consists of over 1,700 psychotherapy sessions conducted as part of six independent clinical trials.  
All of the six trials focused of motivational interviewing for addressing various forms of addiction, including alcohol (ARC, ESPSB, ESB21, CTT), marijuana (iCHAMP), and poly-drug abuse (HMCBI) \cite{atkins2014scaling, baer2009agency}.  
The CTT and HMCBI consist of both real and standardized patients.  
Standardized patients are actors portraying patients with relevant addiction issues for the purpose of therapist training.
A subset of the sessions (N=345) were manually transcribed and subsequently segmented at the utterance level.
The utterances then received behavioral codes according to the Motivational Interviewing Skill Code (MISC) manual.
Of the 345, eight were discarded due to errors or inconsistencies in transcription or behavioral coding, resulting in 337 sessions being considered for the present work.

The MISC manual defines 28 utterance-level behaviors defined in the manual: 19 therapist and 9 client.
We follow the procedure of Xiao et al. \cite{xiao2016behavioral} and group the most infrequent of these codes into composite groups resulting in 11 target labels: 8 therapist and 3 client.
The non-grouped therapist codes are facilitate (FA), giving information (GI), close and open questions (QUC/QUO), simple and complex reflections (RES/REC).
The remainder of the codes are grouped into MI adherent (MIA), i.e., behaviors that adhere to the spirit of the MI treatment, and MI non-adherent (MIN), those which are inconsistent with MI.
Client codes are follow/neutral (FN), which covers the majority of the client utterances.
This code indicates that the client made a statement that was neutral towards changing the targeted behavior of the therapy.
The remainder of the client codes are grouped into positive `change talk' (POS) or negative `sustain talk' (NEG) behaviors.
Change talk is a statement that reflects a client's reasoning, commitments, or steps towards behavior change.
Sustain talk reflects the opposite.
We show the individual and grouped utterance-level MISC codes and their occurrences in the MI dataset in table \ref{T:misc}.

\begin{table}[h]
\caption{\label{T:misc} {MISC code grouping and counts in the dataset.}}
\vspace{2mm}
\centerline{
\begin{tabular}{|llr|}
\hline
Group     & MISC Code & Count \\ \hline \hline
\multicolumn{3}{|c|}{Counselor} \\ \hline \hline
FA       & Facilitate & 14659 \\ \hline
GI       & Giving information & 11880 \\ \hline
QUC      & Closed question    & 6850 \\ \hline
QUO      & Open question      & 5602 \\ \hline
REC      & Complex reflection & 5825 \\ \hline
RES      & Simple reflection  & 8508 \\ \hline
\multirow{4}{*}{MIA}      & MI adherent: Affirm; Reframe; & \multirow{4}{*}{5072} \\ 
         & Emphasize control; Support; Filler;   &  \\
         & Advice with permission; Structure; & \\
         & Raise concern with permission & \\ \hline
\multirow{3}{*}{MIN}      & MI non-adherent: Confront; Direct; & \multirow{3}{*}{1164}\\
         & Advice without permission; Warn; &\\
         & Raise concern without permission & \\ \hline \hline
\multicolumn{3}{|c|}{Client} \\ \hline \hline
FN       & Follow/Neutral     & 37937 \\ \hline
\multirow{3}{*}{POS}      & Change talk: positive  & \multirow{3}{*}{5681} \\
         & Reasons; Commitments; & \\
         & Taking steps; Other & \\ \hline
\multirow{2}{*}{NEG} & Sustain talk: negative & \multirow{2}{*}{4665} \\ 
         & Reasons; Commitments; & \\
         & Taking steps; Other & \\ \hline
\end{tabular}
}
\end{table}


\subsection{Cognitive Behavioral Therapy Corpus}

The Cognitive Behavioral Therapy Corpus consists of over 5,000 audio recordings of therapists conducting cognitive behavioral therapy sessions \cite{creed2016implementation}.  More than 2,000 of these sessions have received manual behavioral coding according to the Cognitive Therapy Rating System (CTRS) manual \cite{young1980cognitive}.
The CTRS defines 11 session-level behavioral codes, which are each scored on a 7 point likert scale (0 `poor' to 6 'excellent').
We pose this as a binary prediction task by assigning codes greater or equal to 3 as `high' and those less than 3 as `low'.
These are given in table \ref{T:ctrs} with the ratio of `high' to `low' labels for each behavioral code.

\begin{table}[!htb]
\caption{\label{T:ctrs} Session-level behavior codes defined by the CTRS manual}
\begin{center}
\begin{tabular}{|clr|} \hline
Abbr. & CTRS Code & Count (`high'/`low') \\ \hline\hline
AG & agenda & 47/45 \\
\multirow{2}{*}{AT} & application of cognitive- & \multirow{2}{*}{44/48} \\
& behavioral techniques & \\
CO & collaboration & 62/30 \\
FB & feedback & 46/46 \\
GD & guided discovery & 48/44 \\
HW & homework & 43/49 \\
IP & interpersonal effectiveness & 82/10 \\
\multirow{2}{*}{KC} & focusing on key cognitions & \multirow{2}{*}{48/44} \\
& and behaviors & \\
PT & pacing and efficient use of time & 51/41 \\
SC & strategy for change & 46/46 \\
UN & understanding & 71/21 \\
\hline
\end{tabular}
\end{center}
\label{default}
\end{table}%

All the defined codes reflect therapist behaviors in the session, there are no codes which reflect client behaviors.
They can be associated into a few broad categories including management and structure of the session (AG, FB, PT, HW), the aspects of the therapist-client relationship (IP, CO, UN), and conceptualization of the clients' presented concerns and approaches for addressing them (GD, KC, SC, AT) \cite{flemotomos2018language}.

Of the behaviorally coded sessions 100 were chosen for manual transcription.
The sessions chosen for transcription were the sessions which received the 50 highest and 50 lowest total ratings (sum across 11 behavioral code ratings).
Eight of these sessions are not considered due to formatting/transcription quality issues, leaving 92 sessions included 70 therapists to be considered in this work.
An initial effort for evaluating the efficacy of using speech and language processing and machine learning to automatically predict these codes has been recently submitted \cite{flemotomos2018language}.

In table \ref{T:micbtcounts} we show counts for the number of sessions, turns, and words in the training and testing splits for each dataset.

\begin{table}[!htb]
\caption{\label{T:micbtcounts} Data Overview: Session, turn, and word counts in training/testing splits.}
\vspace{2mm}
\centerline{
\begin{tabular}{|cccc|}
\hline
Subject & Sessions & Turns & Words \\ \hline \hline
\multicolumn{4}{|c|}{MI} \\ \hline \hline
Counselor & 228/109 & 28.7K/13.9K & 579K/248K  \\
Client & 228/109  & 28.6K/13.6K  & 563K/269K  \\ \hline \hline
\multicolumn{4}{|c|}{CBT} \\ \hline \hline
Counselor & 62/30 & 11.3k/4.5k & 180k/77.4k  \\
Client & 62/30 & 11.5k/4.7k & 215k/109k  \\
\hline
\end{tabular}}
\end{table}

\section{Results and Discussion}

\subsection{Experiments}

All models are implemented in Keras \cite{chollet2015keras} with Theano as the backend \cite{bastien2012theano}.
An early stopping procedure is used in which training is terminated if loss on the validation set does not improve after consecutive epochs.
The validation set is 10\% of the data from the training set which is chosen randomly.
All models are optimized using the adam algorithm with a learning rate of $10^{-3}$ \cite{kingma2014adam}.
As a metric of comparison we use f1-score (macro average across labels).
The reported results are the mean f1-score of the network being trained from 10 different random initializations.

Word embeddings vectors (300 dimensional) are pre-trained on the training data using word2vec \cite{mikolov2013efficient}.
All hidden layers are of the same dimension as the word vectors (initialized with Glorot uniform).
The data is separated into batches of 32 samples and shuffled randomly between epochs.

As described in section \ref{S:mtlmethod} we use an adversarial approach, with the hyper-parameters $\lambda=0.05$ and $\gamma=0.01$ as recommended in \cite{liu2017adversarial} for multi-task learning.
The unshared encoding layers are initialized with the weights from the multi-label systems trained in the previous experiments.
The learning rate of the optimizer is reduced to $10^{-4}$ to allow for fine tuning of these layers.

\begin{table*}[htp]
\caption{Comparison of single-label, multi-label, and multi-label multi-task systems}
\begin{center}
\begin{tabular}{|l|c|cc|cc|c|}
\hline
code & baseline & SL & SL-sw & ML & ML-sw & ML-MT\\
\hline \hline
\multicolumn{7}{|c|}{MISC} \\ \hline \hline
FA & 0.289 & 0.909 & 0.887 & \bf{0.919} & 0.903 & 0.911\\
GI & 0.264 & 0.757 & 0.709 & \bf{0.771} & 0.743 & 0.760\\
QUC  & 0.156 & \bf{0.672} & 0.586 & 0.625 & 0.598 & 0.659\\ 
QUO & 0.122 & \bf{0.802} & 0.639 & 0.798 & 0.787 & 0.801\\
REC & 0.143 & 0.498 & 0.484 & 0.522 & 0.504 & \bf{0.564}\\
RES & 0.185 & 0.476 & \bf{0.516} & 0.491 & 0.429 & 0.486\\
MIA & 0.118 & 0.556 & 0.442 & 0.517 & 0.548 & \bf{0.576}\\
MIN & 0.018 & 0.001 & 0.112 & 0.066 & 0.199 & \bf{0.235} \\
FN & 0.637 & 0.960 & 0.963 & \bf{0.964} & 0.949 & 0.958\\
POS & 0.117 & 0.286 & 0.363 & 0.316 & 0.379 & \bf{0.381}\\
NEG & 0.094 & 0.185 & 0.322 & 0.252 & 0.339 & \bf{0.354} \\ \hline
AVG & 0.195 & 0.555 & 0.548 & 0.567 & 0.580 & \bf{0.608}\\ \hline \hline
\multicolumn{7}{|c|}{CTRS} \\ \hline \hline
AG & 0.667 & 0.718 & 0.718 & 0.716 & 0.784 & \bf{0.790}\\
AT & 0.605 & 0.654 & 0.654 & 0.654 & 0.714 & \bf{0.731}\\
CO & 0.776 & 0.327 & 0.776 & 0.776 & \bf{0.778} & 0.776 \\
FB & 0.636 & 0.686 & 0.686 & 0.687 & 0.751 & \bf{0.772} \\
GD & 0.636 & 0.672 & 0.672 & 0.667 & 0.693 & \bf{0.752} \\
HW & 0.605 & 0.671 & 0.672 & 0.661 & \bf{0.743} & 0.654 \\
IP & 0.929 & 0.000 & 0.929 & 0.929 & 0.929 & 0.929  \\
KC & 0.667 & 0.692 & 0.692 & 0.687 & 0.717 & \bf{0.753} \\
PT & 0.696 & 0.705 & 0.705 & 0.720 & 0.741 & \bf{0.798} \\
SC & 0.605 & 0.644 & 0.644 & 0.642 & 0.695 & \bf{0.744} \\
UN & 0.800 & 0.071 & 0.800 & 0.800 & 0.800 & 0.800 \\ \hline
AVG & 0.688 &0.531 &  0.723 & 0.722 & 0.758 &  \bf{0.773} \\
\hline
\end{tabular}
\end{center}
\label{T:multilabel_mulitask}
\end{table*}%

\subsubsection{Multi-label Learning}

In this section we compare single-label (SL) and multi-label (ML) approaches with and without sample weighting (sw) for predicting behavioral codes in our exemplary domains.  
In table \ref{T:multilabel_mulitask}, we show prediction results (f1-score) for the MISC and CTRS behavioral code prediction tasks.
As a point of reference, we include the f1-score for each behavioral code if that behavior is considered present in every turn referred to as `baseline' in the table. 
Due to the imbalance in the labels in both datasets, sample weights improve performance for both single-label and multi-label prediction.

For MISC code prediction, the multi-label approaches outperform single-label approaches both with and without sample weighting.
The highest per-code results are for multi-label without sample weights for the codes FA, GI, QUC, QUO, REC, RES, and FN.
The occurrence of these codes is more balanced thus they do not benefit from the sample weighting scheme.
However, the more unbalanced codes (MIA, MIN, POS, and NEG) are predicted best by the multi-label system with sample weights, due to the infrequency in which they occur in the data.
On average, the ML-sw approach resulted in the highest performance for the MISC prediction task.

With respect to the CTRS behavioral code prediction task, the multi-label system with sample weighting yielded the highest (or tied) f1-score performance for all CTRS codes.
In this case, however, the mutli-label system without sample weighting did not outperform the single-label approach using sample weights.

\subsubsection{Multi-task Learning}

In this section we evaluate the performance of a multi-task model that aims to learn both MISC and CTRS behavioral codes.  
We show the results for the multi-task model in table \ref{T:multilabel_mulitask}.  
The average performance of both tasks is improved compared to their single-task counterparts.

In the MISC task, prediction of MIN (MI non-adherent) had the largest relative improvement (18.1\%) with respect to the best result from the single-task approaches.
MI non-adherent behaviors include confrontation, direction, warning, advising and raising concern without permission.
These are considered negative therapist behaviors in motivation interviewing and are to be avoided by therapists practicing MI due to its non-confrontational and non-adversarial nature.
For this reason, their occurrence is very rare in the MI corpus.
The combined occurrences of these five behaviors combined account for only 2.3\% of all therapist turns in the dataset.
Cognitive behavior therapy has a distinctly different approach with respect to directive statements.
This is evidenced by the behavior of assigning homework (HW), an essential element of CBT.
Thus statements that would be considered non-adherent in MI counseling are much more likely to occur in CBT sessions.
In fact, turns from the CBT dataset are predicted to be MI non-adherent at more than twice the rate than turns from the MI dataset (16.6\% versus 7.4\%). 
Analysis of these turns suggests some of these utterances contain behaviors that are considered counter to the aims of both therapies (i.e., low empathy statements) for example, 
\begin{quote} 
therapist: so i need you to tell me what to do to help you. 
\end{quote}
While, many of the utterances in the CBT data that are labeled as MI non-adherent do not conflict with the spirit of cognitive behavioral therapy (e.g., directive statements) such as, 
\begin{quote} 
therapist: see if there's a way for you to kind of challenge the belief,
\end{quote}
 but do not adhere to the MI counseling style.
In this way the multi-task system allows for a better representation of these behaviors even though the turns in the CBT dataset are not labeled for these behaviors.

With respect to the CTRS task, guided discovery (GD) had the largest relative improvements (8.5\%) for the multi-task system with respect to the single-task system.
Guided discovery is a behavior in which the therapist ``uses exploration and questioning to help patients see new perspectives."
In the MI data, questions are explicitly labeled and therefore encoded by the system.
Because of the important relation between questioning and GD, the multi-task system enables a better representation for decoding this behavior, despite not having turns manually labeled as questions.
The multi-task system predicted turns from CBT sessions with high guided discovery scores to be questions (open or closed) 18.3\% versus 13.5\% from sessions labeled as low guided discovery.


\begin{table*}[htb]
\caption{Multi-label Learning with Context}
\begin{center}
\begin{tabular}{|l|ccccc|ccccc|}
\hline
$C$ & 0 & 1 & 2 & 3 & 4 & 0 & 1 & 2 & 3 & 4 \\
\hline \hline
& \multicolumn{5}{c|}{ML} & \multicolumn{5}{c|}{ML-MT} \\
\hline \hline
\multicolumn{11}{|c|}{MISC} \\
\hline\hline
FA & 0.903 & 0.918 & 0.912 & 0.918 & 0.917 & 0.911 & 0.918 & 0.918 & 0.917 & \bf{0.919} \\ 
GI & 0.743 & 0.762 & 0.756 & 0.770 & 0.764 & 0.760 & 0.775 & 0.774 & 0.761 & \bf{0.776}\\ 
QUC & 0.598 & 0.648 & 0.634 & 0.653 & 0.659 & 0.659 & 0.656 & 0.667 & 0.638 & \bf{0.686}\\ 
QUO & 0.787 & 0.803 & 0.801 & 0.809 & 0.809 & 0.801 & 0.809 &\bf{0.812} & \bf{0.812} & 0.806\\ 
REC & 0.504 & 0.549 & 0.560 & 0.558 & \bf{0.594} & 0.564 & 0.592 & 0.576 & 0.572 & 0.570 \\  
RES & 0.429 & 0.463 & 0.461 & 0.495 & 0.504 & 0.486 & \bf{0.519} & 0.504 & 0.499 & 0.516 \\ 
MIA & 0.548 & 0.565 & 0.532 & 0.570 & 0.558 & 0.576 & 0.580 & \bf{0.587} & 0.551 & 0.581 \\ 
MIN & 0.199 & 0.213 & 0.191 & 0.224 & 0.220 & \bf{0.235} & 0.223 & 0.221 & 0.229 & 0.208 \\ 
FN & 0.949 & 0.956 & 0.956 & \bf{0.960} & 0.954 & 0.958 & 0.959 & 0.956 & \bf{0.960} & \bf{0.960} \\ 
POS & 0.379 & 0.405 & 0.371 & 0.408 & 0.401 & 0.381 & 0.396 & \bf{0.416} & 0.332 & 0.397 \\ 
NEG & 0.339 & 0.372 & 0.361 & 0.365 & \bf{0.384}  & 0.354 & 0.377 & 0.372 & 0.383 & 0.391 \\ 
\hline
AVG & 0.580 & 0.605 & 0.594 & 0.612 & 0.615 & 0.608 & \bf{0.619} & \bf{0.619} & 0.605 & \bf{0.619} \\ 
\hline \hline
\multicolumn{11}{|c|}{CTRS} \\
\hline\hline
AG & 0.784 & 0.771 & 0.787 & 0.732 & 0.766 & \bf{0.790} & 0.739 & 0.772 &  0.771 & 0.741 \\ 
AT & 0.714 & 0.733 & \bf{0.749} & 0.691 & 0.739 & 0.731 & 0.712 & 0.739 &  0.742 & 0.707 \\ 
CO & 0.778 & 0.787 & \bf{0.792} & 0.790 & 0.789 & 0.776 & 0.775 & 0.777 &  0.783 & 0.774 \\ 
FB & 0.751 & 0.750 & 0.770 & 0.716 & 0.754 & 0.772 & 0.753 & \bf{0.778} &  0.753 & 0.712 \\ 
GD & 0.693 & 0.741 & 0.772 & 0.758 & \bf{0.780} & 0.752 & 0.771 & 0.770 &  0.746 & 0.764 \\ 
HW & \bf{0.743} & 0.705 & 0.731 & 0.637 & 0.737 & 0.654 & 0.643 & 0.723 &  0.735 & 0.703 \\ 
IP & 0.929 & 0.929 & 0.929 & 0.929 & 0.929 & 0.929 & 0.929 & 0.929 &  0.929 & 0.929 \\ 
KC & 0.717 & 0.736 & \bf{0.765} & 0.722 & 0.743 & 0.753 & 0.753 & 0.757 &  0.726 & 0.726 \\ 
PT & 0.741 & 0.767 & 0.779 & 0.780 & 0.792 & 0.798 & 0.797 & \bf{0.828} &  0.800 & 0.794 \\ 
SC & 0.695 & 0.726 & 0.746 & 0.695 & 0.743 & 0.744 & 0.737 & \bf{0.752} &  0.702 & 0.718 \\ 
UN & 0.800 & 0.800 & 0.800 & 0.800 & \bf{0.803} & 0.800 & 0.800 & 0.800 &  0.800 & 0.800 \\ \hline
AVG & 0.758 &   0.768  &  \bf{0.784}  &  0.750  &  0.780 & 0.773 & 0.765 & \bf{0.784} &  0.772 & 0.761 \\ 
\hline
\end{tabular}
\end{center}
\label{T:mulilabel_context_mi_cbt}
\end{table*}%

\subsubsection{Learning with Turn Context}

In table \ref{T:mulilabel_context_mi_cbt}, we present results comparing prediction performance when adding context to the multi-label prediction task.
Adding contextual information provides increased performance for almost every behavioral code.
In the MISC task, the codes REC and RES (complex and simple reflections) had the largest relative improvement (17.9\% and 17.5\% respectively).
Reflections are when a therapist restates information provided by the client and are either a slight rephrasing (simple) or add significant meaning or emphasis (complex) to the client's statements.
Thus, it is intuitive that prediction of these behaviors would benefit from knowing the surrounding utterances.
One queue that indicates a simple reflection is a therapist repeating verbatim what was stated by the client, for example,
\begin{quote}
therapist: you're sober how many days? \\
client: just like thirty five days. \\
therapist: thirty five days sober.
\end{quote}
Clearly, having the context of nearby utterances enables better prediction of such occurrences.

In the CTRS task, the codes guided discovery (GD), strategy for change (SC), and pacing and timing (PT), had the most relative performance improvement with context (12.6\%, 7.3\%, and 6.9\% respectively).  
Guided discovery and strategy for change are both behaviors which reflect the therapists' conceptualization of the client's concern and their approach for addressing them.  
Therefore, these are behaviors that unfold and occur throughout the session not in isolation.
Pacing and timing (PT) reflects the therapist's ability to manage the pace of the session over the course of the session and thus turn context will provide useful information about the therapist's skill in this regard.

In the CTRS task, the codes agenda (AG) and homework (HW) did not improve with added context.  These codes typically only comprise a small portion of the session (one or two turns) as agenda simply establishes what will be discussed and homework refers to tasks the therapist will assign the client at the end of the session to perform before the next session, thus turn context is not helpful in these scenarios.

\subsubsection{Multi-label multi-task learning with context}

We show the results for the a system combining the multi-label and multi-task paradigms with context in table \ref{T:mulilabel_context_mi_cbt}.
The mutli-label, multi-task system with context achieved the highest combined performance for the two tasks.  
The average performance for the CTRS task did not improve in this setting versus the single-task multi-label system with context.  
The CTRS prediction loss typically converges more quickly than that of the MISC, likely due in part to the amount of available data as well as the level of supervision (session labels versus turn labels). 
The multi-label multi-task system achieved higher performance than the single task system for 8 of 11 MISC behavioral codes and 4 of 11 CTRS behavioral codes.
 The CTRS behavioral codes interpersonal effective (IP) and understanding (UN) had the most extreme label imbalance and thus only UN achieved performance above the baseline in the case of multi-label single-task with context of 4 turns.

\section{Conclusions}
%

In this work we proposed multi-label and multi-task approaches for behavioral coding of psychotherapy interactions.  
We demonstrated that by incorporating these paradigms which help reflect the complexities of these data better prediction of behaviors in these sessions is achieved. 
Multi-label learning benefited prediction of less frequently occurring behaviors by learning a model that takes advantages of a representation that incorporates modeling of more frequent behaviors allowing for a richer representation.
Multi-task learning benefited prediction of codes in both corpora by taking advantage of a model that incorporates behaviors which are common among the datasets.
Using a model that incorporates turn context improved prediction of most behaviors in both tasks by incorporating more relevant information from the session.
 The multi-label multi-task system with turn context achieved the highest combined prediction for the behavioral coding tasks.
Additionally, we discussed the particular behaviors which yielded the highest prediction performance improvement using the proposed methodology.

\subsection{Applications of Automatic Behavioral Coding}
Providing automated methods for coding behaviors which occur in psychotherapy interactions has many potential applications.  
One of the first proposed applications is the task of evaluating therapist efficacy from therapy audio recordings using a speech pipeline system which performs audio segmentation, automatic transcription, and behavioral code prediction \cite{xiao2015rate}.
Such a framework could enable patients to choose their therapist based on empirically derived quality metrics rather than word of mouth and online reviews.
Additionally, this could enable monitoring patient progress and tracking of behavioral changes and symptoms over time.
Automatic behavioral coding has the potential to augment therapist understanding of their clients and the quality of the therapy they are providing by allowing rapid monitoring and feedback of their therapy sessions.
Furthermore, lessons learned from ABC developments within the psychotherapy domain may provide insights to automatic understanding and modeling of human behaviors in other human-human and human-computer interaction domains.

\subsection{Future Work}

There are many potential avenues for extending the proposed work.
One key direction is to investigate how these learning paradigms are affected by imperfect word, speaker, and turn boundary information that would be derived from a speech pipeline system \cite{xiao2016technology}.
Such an investigation is necessary to determine the feasibility of incorporating complex learning paradigms in a truly automatic behavioral coding system.

While the present work did not incorporate multi-modal feature representations, it is an important line of inquiry.
As discussed in section \ref{S:mlforbc}, there are many behavioral cues which are important for the behavioral coding task.
One such effort \cite{singla2018using}, proposes fusing lexical and prosodic information in an attentional LSTM to predict behaviors in MI therapy sessions.
The promising results of this initial step encourage further exploration of this area.

In addition to evaluating alternative feature representations, we are interested in combining aspects of the proposed methodology with other deep learning approaches such as hierarchical attention networks \cite{yang2016hierarchical} and domain adaptation networks \cite{glorot2011domain}.
Hierarchical attention networks provide attention weighting for learning turn context, which may allow for a contextual system to only focus on the most relevant turns in the conversation.
Domain adaptation networks learn representations from data of one domain and then adapt the representation to data of a target domain.
This approach is a type of inductive transfer learning, where data from domains that are readily available can be used to augment learning for domains where data is harder to collect.
This could be of special interest in psychotherapy and behavioral health domains where data are often of an extremely private and sensitive nature.


\bibliographystyle{IEEEtran}
\bibliography{allcitations.bib}

\vfill


\end{document}